\documentclass[sigconf]{acmart}
\usepackage{hyperref}
\usepackage{url}
\usepackage{microtype}
\usepackage{caption}
\usepackage{subcaption}
\usepackage{booktabs} 
\usepackage{amsmath}
\usepackage{hyperref}
\usepackage{cleveref}
\usepackage{tabularx}

\newcommand{\divergence}[1]{D_{\textnormal{#1}}}
\DeclareMathOperator{\IS}{IS}
\DeclareMathOperator{\IN}{IN}
\AtBeginDocument{%
  \providecommand\BibTeX{{%
    \normalfont B\kern-0.5em{\scshape i\kern-0.25em b}\kern-0.8em\TeX}}}

\copyrightyear{2021}
\acmYear{2021}
\setcopyright{rightsretained}
\acmConference[KDD '21]{Proceedings of the 27th ACM SIGKDD Conference on Knowledge Discovery and Data Mining}{August 14--18, 2021}{Virtual Event, Singapore}
\acmBooktitle{Proceedings of the 27th ACM SIGKDD Conference on Knowledge Discovery and Data Mining (KDD '21), August 14--18, 2021, Virtual Event, Singapore}
\acmDOI{10.1145/3447548.3467198} 
\acmISBN{978-1-4503-8332-5/21/08}

\begin{document}
\fancyhead{}
\title{On Training Sample Memorization: Lessons from Benchmarking Generative Modeling with a Large-scale Competition}

\author{Ching-Yuan Bai}
\email{b05502055@csie.ntu.edu.tw}
\affiliation{%
  \institution{
  Computer Science and Information Engineering\\
National Taiwan University}
  \country{Taiwan}
}

\author{Hsuan-Tien Lin}
\email{htlin@csie.ntu.edu.tw}
\affiliation{%
  \institution{Computer Science and Information Engineering\\
National Taiwan University}
  \country{Taiwan}
}

\author{Colin Raffel}
\email{craffel@google.com}
\affiliation{%
  \institution{Google}
  \country{USA}
}

\author{Wendy Chih-wen Kan}
\email{wendykan@google.com}
\affiliation{%
  \institution{Kaggle, Google}
  \country{USA}
}
\renewcommand{\shortauthors}{Bai et al.}

\begin{abstract}
Many recent developments on generative models for natural images have relied on heuristically-motivated metrics that can be easily gamed by memorizing a small sample from the true distribution or training a model directly to improve the metric.
In this work, we critically evaluate the gameability of these metrics by designing and deploying a generative modeling competition.
Our competition received over 11000 submitted models. 
The competitiveness between participants allowed us to investigate both intentional and unintentional memorization in generative modeling. 
To detect intentional memorization, we propose the ``Memorization-Informed Fr\'echet Inception Distance'' (MiFID) as a new memorization-aware metric and design benchmark procedures to ensure that winning submissions made genuine improvements in perceptual quality.
Furthermore, we manually inspect the code for the 1000 top-performing models to understand and label different forms of memorization.
Our analysis reveals that unintentional memorization is a serious and common issue in popular generative models.
The generated images and our memorization labels of those models as well as code to compute MiFID are released to facilitate future studies on benchmarking generative models.
\end{abstract}




\maketitle

\section{Introduction}
Recent work on generative models for natural images has produced huge improvements in image quality, with some models producing samples that can be indistinguishable from real images \cite{karras2017progressive,karras2019style,karras2019analyzing,brock2018large,kingma2018glow,maaloe2019biva,menick2018generating,razavi2019generating}.
Improved sample quality is important for tasks like super-resolution \cite{ledig2017photo} and inpainting \cite{yu2019free}, as well as creative applications \cite{park2019semantic,isola2017image,zhu2017unpaired,zhu2017your}.
These developments have also led to useful algorithmic advances on other downstream tasks such as semi-supervised learning \cite{kingma2014semi,odena2016semi,salimans2016improved,izmailov2019semi} or representation learning \cite{dumoulin2016adversarially,donahue2016adversarial,donahue2019large}.

Modern generative models utilize a variety of underlying frameworks, including autoregressive models \cite{oord2016pixel}, Generative Adversarial Networks \citep[GANs;][]{goodfellow2014generative}, flow-based models \cite{dinh2014nice,rezende2015variational}, and Variational Autoencoders \citep[VAEs;][]{kingma2013auto,rezende2014stochastic}.
This diversity of approaches, combined with the subjective nature of evaluating generation performance, has prompted the development of heuristically-motivated metrics designed to measure the perceptual quality of generated samples such as the Inception Score~\citep[IS;][]{salimans2016improved} or the Fr\'echet Inception Distance~\citep[FID;][]{heusel2017gans}.
These metrics are used in a benchmarking procedure where ``state-of-the-art'' results are claimed based on a better score on standard datasets.

Indeed, much recent progress in the field of machine learning as a whole has relied on useful benchmarks on which researchers can compare results.
Specifically, improvements on the benchmark metric should reflect improvements towards a useful and nontrivial goal.
Evaluation of the metric should be a straightforward and well-defined procedure so that results can be reliably compared.
For example, the ImageNet Large-Scale Visual Recognition Challenge \cite{deng2009imagenet,russakovsky2015imagenet} has a useful goal (classify objects in natural images) and a well-defined evaluation procedure (top-1 and top-5 accuracy of the model's predictions).
Sure enough, the ImageNet benchmark has facilitated the development of dramatically better image classification models which have proven to be extremely impactful across a wide variety of applications.

Unfortunately, some of the commonly-used benchmark metrics for generative models of natural images do not satisfy the aforementioned properties.
For instance, although the IS is demonstrated to correlate well with human perceived image quality~\citep{salimans2016improved}, \citet{barratt2018note} point out several flaws of the IS when used as a single metric for evaluating generative modeling performance, including its sensitivity to the choice of a representational space, which undermines generalization capability.
Separately, directly optimizing a model to improve the IS can result in extremely \textit{unrealistic}-looking images~\citep{barratt2018note} despite resulting in a better score.
It is also well-known that a good IS can be achieved~\citep{gulrajani2018towards} by memorizing images from the training set (i.e.\ producing \textit{non-novel} images).
On the other hand, FID is widely accepted as an improvement over IS due to its better consistency under perturbation~\cite{heusel2017gans}.
However, there is no clear evidence of FID resolving any of the flaws of the IS.

Motivated to better understand the potential misalignment between the goal and the benchmark in generative modeling, 
we benchmark generative models and critically examine the metrics used to evaluate them by holding a public machine learning competition.
To the extent of our knowledge, no large-scale generative modeling competitions have ever been held, possibly due to the immense difficulty of measuring perceptual quality and identifying training sample memorization in an efficient and scalable manner.
We modified FID to autonomously penalize competition submissions with
memorization in an attempt to discourage contestants from intentionally memorizing training images.
We also manually inspected the code for the top 1000 submissions to reveal different forms of intentional or unintentional memorization, to ensure that the winning submissions reflect meaningful improvements, and to confirm efficacy of our proposed metric.
We hope that the success of the first-ever generative modeling competition can serve as future reference and stimulate more research in developing better generative modeling benchmarks.

The remainder of this paper is structured as follows:
In Section~\ref{sec:background}, we briefly review the metrics and challenges of evaluating generative models.
In Section~\ref{sec:design}, we explain in detail the competition design choices and propose a novel benchmarking metric, Memorization-Informed Fréchet Inception Distance (MiFID).
We show that MiFID enables fast profiling of participants that intentionally memorize the training examples.
In Section~\ref{sec:data-release}, we introduce a dataset released along with this paper that includes over one hundred million generated images and labels of the memorization methods adopted in the generation process (derived from manual code review).
In Section~\ref{sec:insights}, we connect the phenomena observed in large-scale benchmarking of generative models in the real world back to the research community and point out crucial but neglected flaws in FID.

\label{sec:background}
In generative modeling, the goal is to produce a model $p_\theta(x)$, parameterized by $\theta$, that approximates some true distribution $p(x)$.
We are not given direct access to $p(x)$; instead, we are provided only with samples $x$ drawn from $p(x)$.
In this paper, we will assume that samples $x$ from $p(x)$ are $64$-by-$64$ pixel natural images, i.e.\ $x \in \mathbb{R}^{64 \times 64 \times 3}$.
A common approach is to optimize $\theta$ so that $p_\theta(x)$ assigns high likelihood to training examples drawn from $p(x)$.
This provides a natural evaluation procedure which measures the likelihood assigned by $p_\theta(x)$ to held-out examples drawn from $p(x)$.
However, not all generative models returns an explicit form of $p_\theta(x)$ that can be directly computed.

Notably, Generative Adversarial Networks (GANs) \cite{goodfellow2014generative} learn an ``implicit'' model that can be sampled from but do not provide an exact (nor even an estimate of) $p_\theta(x)$ for any given $x$.
The GANs have proven particularly successful at producing models that can generate extremely realistic and high-resolution images, which leads to a natural question: How should we evaluate the quality of a generative model if we cannot compute the likelihood $p_\theta$ assigned to held-out examples (following the conventional  approach of evaluating generalization in machine learning)?

This question has led to the development of many alternative ways to evaluate generative models \cite{borji2019pros}.
A historically popular metric, proposed in \cite{salimans2016improved}, is the Inception Score (IS) which computes
\[
        \IS(p_\theta) = \mathbb{E}_{x \sim p_\theta(x)} [ \divergence{KL}(\IN(y | x) \| \IN(y)) ]
\]
where $\IN(y | x)$ is the conditional probability of a class label $y$ assigned to a datapoint $x$ by a pre-trained Inception Network \cite{szegedy2015going}.
More recently, \citep{heusel2017gans} proposed the Fr\'echet Inception Distance (FID) which is claimed to better correlate with perceptual quality.
FID uses the estimated mean and covariance of the Inception Network feature space distribution to calculate the distance between the real and fake distributions up to the second order.
FID between the real images $r$ and generated images $g$ is computed as:
\[
    \text{FID}(r, g)=\left\|\mu_{r}-\mu_{g}\right\|_{2}^{2}+\operatorname{Tr}\left(\Sigma_{r}+\Sigma_{g}-2\left(\Sigma_{r} \Sigma_{r}\right)^{\frac{1}{2}}\right)
\]
where $\mu_{r}$ and $\mu_{g}$ are the mean of the real and generated images in latent space, and $\Sigma_{r}$ and $\Sigma_{g}$ are the covariance matrices for the real and generated feature vectors.
A drawback of both IS and FID is that they assign a very good score to a model which simply memorizes a small and finite sample from $p(x)$ \cite{gulrajani2018towards}, an issue we address in Section \ref{sec:mifid}.

\section{Generative Modeling Competition Design}
\label{sec:design}
We designed the first generative model competition,\footnote{\url{https://www.kaggle.com/c/generative-dog-images}} hosted by Kaggle, where participants were invited to generate realistic dog images given $20{,}579$ images of dogs from ImageNet \cite{russakovsky2015imagenet}.
Participants were required to implement their generative models in a constrained computation environment to prevent them from obtaining unfair advantages (e.g.\ downloading additional dog images).
The computation environment was designed with:
\begin{itemize}
    \item Limited computation resource (9 hours on a NVIDIA P100 GPU for each submission) since generative model performance is known to be highly related to the amount of computational resources used~\cite{brock2018large}.
    This allows us to remove the factor of computation budget when analyzing the results.
    \item Isolated containerization to avoid continuous training by reloading model checkpoints from previous sessions.
    \item No access to external resources (i.e.\ the internet) to avoid usage of pre-trained models or additional data. 
\end{itemize}
Each submission is required to provide $10{,}000$ generated images of dimension $64 \times 64 \times 3$ and receives a public score in return. 
Participants are allowed to submit any number of submissions during the two-month competition.
Before the end of the competition, each team is required to choose two submissions, and the final ranking is determined by the better private score (described below) out of the two selected submissions.

Top 5 place winners on the private leaderboard received monetary rewards (\$$2{,}000$ each), while top $100$ winners receive Kaggle medals. 

\begin{figure*}
\centering
  \includegraphics[width=0.8\linewidth]{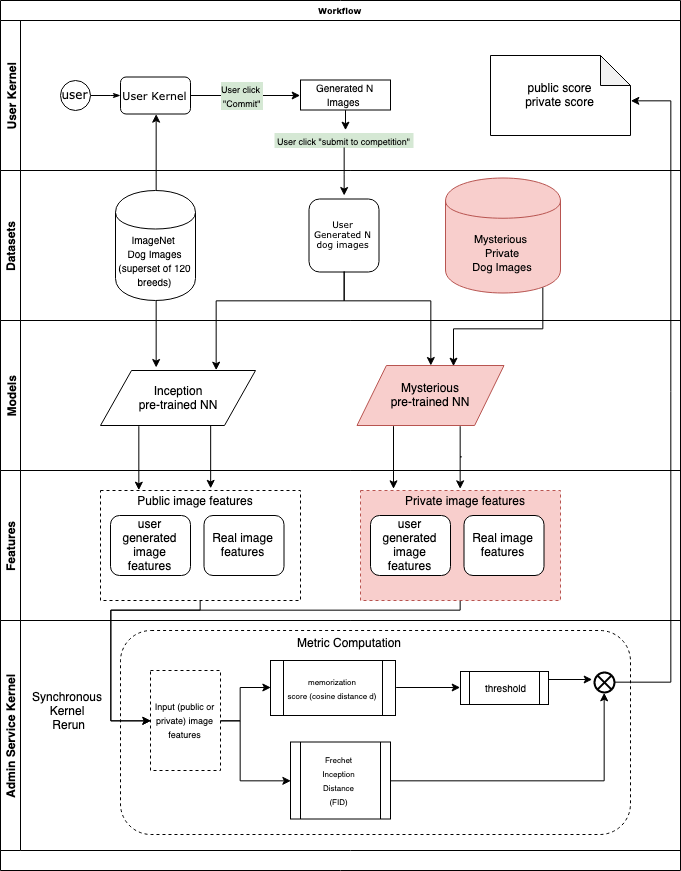}
  \caption{Workflow calculating MiFID for public and private scores. Note that the ``mysterious'' neural network and dataset were kept intentionally vague to prevent score overfitting. The details of them are described in Table \ref{table:public-private} }
  \label{fig:workflow}
\end{figure*}

In the following sections, we discuss how the final decisions were made regarding pretrained model selection (for FID feature projection) and how we enforced penalties to ensure the fairness of the competition. The system design is demonstrated in Figure~\ref{fig:workflow}.

\subsection{Memorization-Informed Fréchet Inception Distance (MiFID)}
\label{sec:mifid}
The most crucial part of the competition is the performance evaluation metric to score the submissions.
To assess the quality of generated images, we adopted the Fréchet Inception Distance~\cite{heusel2017gans} which is a widely used metric for benchmarking GANs.
Compared to the Inception Score~\cite{salimans2016improved}, FID has the benefits of better robustness against noise and distortion and more efficient computation~\cite{borji2019pros}.

For a generative modeling competition, a good metric not only needs to reflect the quality of generated samples but must also allow for easy identification of intentional-memorization with as little manual intervention as possible.
Many forms of intentional-memorization were prevented by setting up the aforementioned computation environment, but even with these safeguards it would be possible to ``game'' FID.
Specifically, we predicted that memorization of training data would be a major issue, since current generative model evaluation metrics such as IS or FID are prone to assign high scores to models that regurgitate memorized training data \cite{gulrajani2018towards}.
This motivated the addition of a "memorization-aware" metric that penalizes models producing images too similar to the training set.

Combining memorization-aware and generation quality components, we introduced Memorization-Informed Fr\'echet Inception Distance (MiFID) as the metric used for the competition:
\[
\text{MiFID}(S_g, S_t) = m_\tau(S_g, S_t) \cdot \text{FID}(S_g, S_t)
\]
where $S_g$ is the generated set, $S_t$ is the original training set, FID is the Fr\'echet Inception Distance, and $m_\tau$ is the memorization penalty, as discussed below.

\subsubsection{Memorization Penalty}
To capture the similarity between two sets of data -- in our case, generated images and original training images -- we started by measuring similarity between individual images.
We opted to use the cosine similarity in a learned representation space to compare images.
The cosine similarity is easy to implement with high computational efficiency (with existing optimized BLAS libraries) which is ideal when running a competition with hundreds of submissions each day.
The value is also bounded, making it possible to intuitively understand and compare the degree of similarity.

We define the memorization distance $s$ of a target projected generated set $S_g \subseteq \mathbb{R}^d$ with respect to a reference projected training set $S_t \subseteq \mathbb{R}^d$ as 1 subtracted by the mean of minimum (signed cosine) similarity of all elements $S_g$ and $S_t$.
Intuitively, lower memorization distance is associated with more severe training sample memorization.
Note that the distance is asymmetric i.e. $s(S_g, S_t) \neq s(S_t, S_g)$, but this is irrelevant for our use-case.
\[
    s(S_g, S_t) := \frac{1}{|S_g|} \sum_{x_g \in S_g} \min_{x_t \in S_t} 
    \bigg{(} 1 -  \frac{|\langle x_g, x_t \rangle|}{|x_g| \cdot |x_t|} \bigg{)}
\]

We hypothesize that submissions with intentional memorization would generate images with significantly lower memorization distance.
To leverage this idea, only submissions with distance lower than a specific threshold $\tau$ are penalized.
Thus, the memorization penalty $m_\tau$ is defined as
\[
    m_\tau(S_g, S_t) = \left\{\begin{array}{ll}
    {\frac{1}{s(S_g, S_t) + \epsilon} \quad (\epsilon \ll 1),} & {\text { if } s(S_g, S_t) < \tau} \\
    {1,} & {\text { otherwise }}
    \end{array}\right.
\]
More memorization (subceeding the predefined threshold $\tau$) will result in higher penalization.
Dealing with false positives and negatives under this penalty scheme is further discussed in Section~\ref{sec:final-ranks}.

\subsubsection{Preventing overfitting}

In order to prevent participants of the competition from overfitting to the public leaderboard, 
we used different data for calculating the public and private score, and generalized FID to a different feature projection space.
Specifically, we selected different pre-trained ImageNet classification models for public and private score calculation.
For each score, we take one pre-trained model, and use the model to compute both the memorization penalty and FID.
Inception V3 was used for the public score following past literature, while NASNet~\cite{zoph2018learning}was used for the private score.
Table \ref{table:public-private} shows the different datasets and models used in public and private leaderboards.
We discuss how NASNet was selected in Section~\ref{sec:model-selection}.

\subsection{Determining Final Ranks}
\label{sec:final-ranks}

After the competition was closed to submission there was a two-week window to re-process all the submissions and remove those that violated the competition rules (e.g.\ by intentionally memorizing the training set) before the final private leaderboard was announced.
The memorization penalty term in MiFID was efficiently configured for re-running with a change of the parameter $\tau$, allowing finalizing of results within a short time frame. 

\subsubsection{Selecting Pre-trained Model for the Private Score}
\label{sec:model-selection}

As it is commonly assumed that FID is generally invariant to the projection space, the pre-trained model for private score was selected to best combat intentional-memorization via training set memorization.
The goal is to separate legitimate and illegitimate submissions as cleanly as possible.
We calculated the memorization distance for a subset of submissions projected with the chosen pre-trained model and coarsely label whether the submission intentionally memorized training samples.
Coarse labeling of submissions was achieved by exploiting competition-related clues.

\begin{figure}
\centering
  \includegraphics[width=0.9\linewidth]{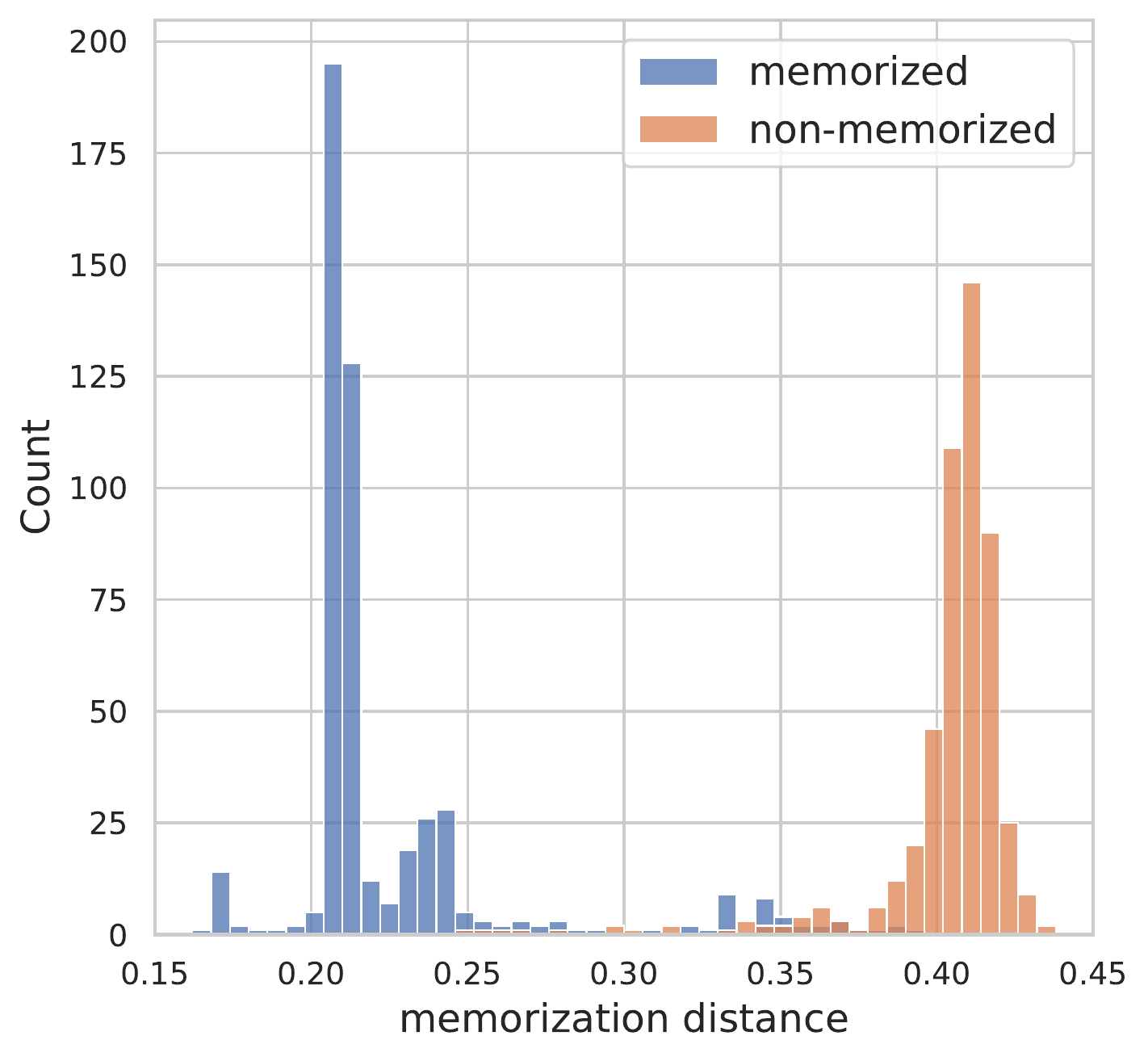}
\caption{Histogram of memorization distance for private leaderboard (using NASNet). The two classes (legitimate models and illegitimate models) are well separated.}
\label{fig:dist-histograms-private}
\end{figure}

\begin{figure}
\centering
\includegraphics[width=0.9\linewidth]{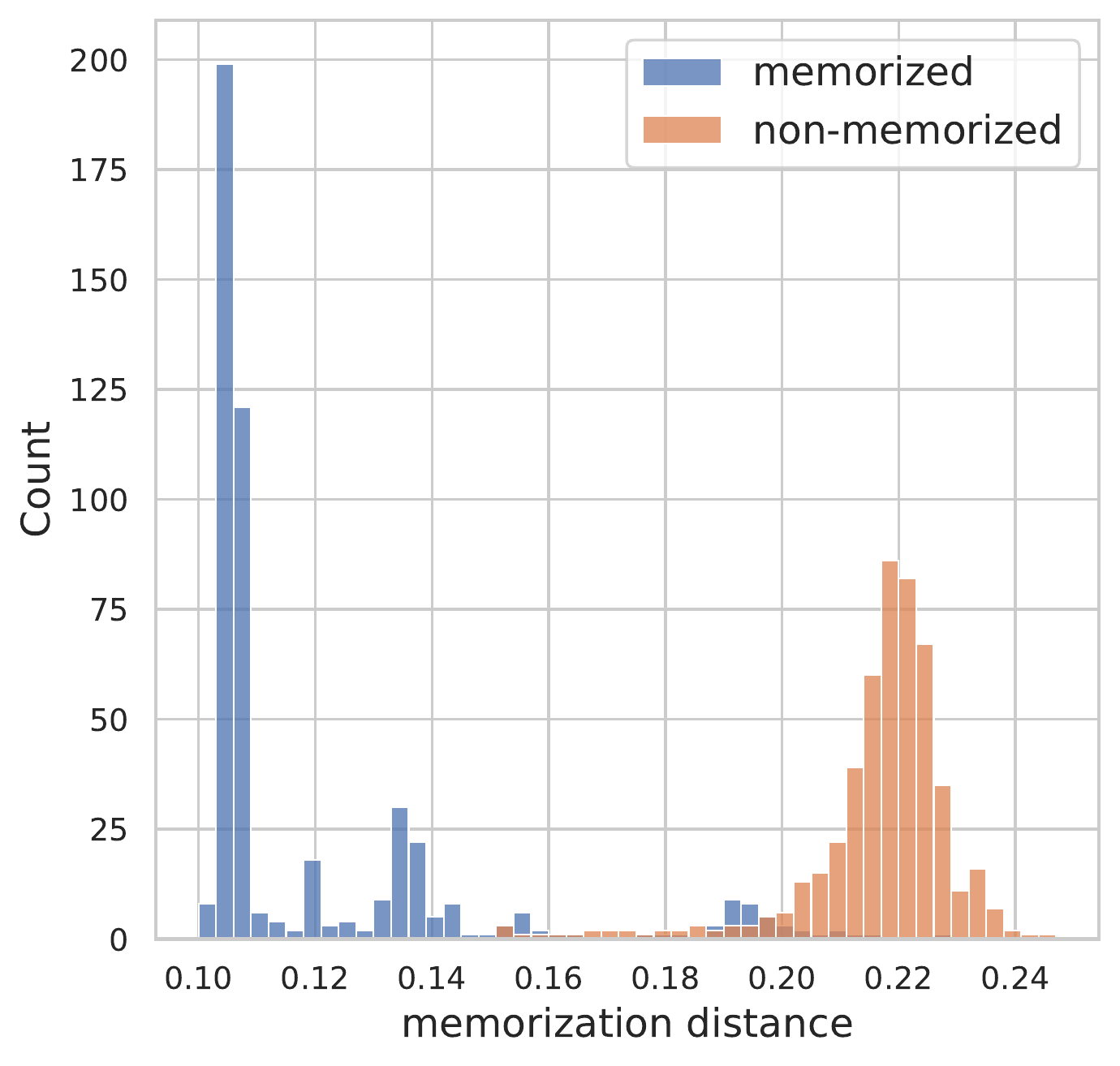}
\caption{Histogram of memorization distance for public leaderboards (using Inception). The two classes (legitimate models and illegitimate models) are well separated.}
\label{fig:dist-histograms-public}
\end{figure}

There exists a threshold $\tau^*$ that best separates memorized versus non-memorized submissions via the memorization distance (see Figure~\ref{fig:dist-histograms-private},~\ref{fig:dist-histograms-public}).
Here we define the memorization margin $d$ of pre-trained model $M$ as
\[
d(M) = \min_{\tau}\sum_{\forall S_g} (s(S_g, S_t) - \tau)^2
\]
The pre-trained model with largest memorization margin was then selected for calculation of the private score, in this case,  NASNet~\cite{zoph2018learning}, and the optimal corresponding memorization penalty $m_\tau$ where $\tau = \tau^*$.
Selecting the threshold after the competition allows the threshold to adapt to the submissions better and the fairness is ensured as any false penalization were well handled.

\begin{table}[ht]
\caption{Configurations for the public and private scores}
\label{table:public-private}
\begin{center}
\begin{small}
\begin{sc}
\begin{tabular}{|l|l|l|}
\hline
               & Public  & Private                                        \\ 
\hline
Model          & Inception & NASNet
                                        
\\ \hline
DataSet        & \begin{tabular}[c]{@{}l@{}}ImageNet dogs \\ 120 breeds, \\ 20579 images\end{tabular} & \begin{tabular}[c]{@{}l@{}}ImageNet dogs +\\ private \\ dogs +\\ Internet dogs\end{tabular} \\ \hline
\end{tabular}
\end{sc}
\end{small}
\end{center}
\end{table}

\subsubsection{Handling False Penalization}
\label{sec:false-penalization}
While MiFID was designed to handle penalization automatically, in practice,  we observed minor mixing of legitimate and illegitimate submissions between the well-separated peaks (Figure~\ref{fig:dist-histograms-private},~\ref{fig:dist-histograms-public}).
While it is well accepted that no model can be perfect, it was necessary to ensure that competition was fair. 
Therefore, different strategies were adopted to resolve false positives and negatives.
For legitimate submissions that were falsely penalized (false positives), participants were allowed to actively submit rebuttals for the result.
For illegitimate submissions that were dissimilar enough to the training set to dodge penalization (false negatives), the team's submitted code was manually reviewed to determine if intentional memorization was present. 
This manual reviewing process of code submissions was labor intensive, as it required expert knowledge of generative modeling, and could not be accomplished by human raters unfamiliar with machine learning. 
The goal was to review enough submissions such that the top 100 teams on the leaderboard would be free of intentional-memorization and others, since we reward the top 100 ranked teams.
Thanks to our design of MiFID, it is possible to set the penalty threshold $\tau$ such that we were comfortable that most users ranked lower than 100 on the leaderboard who intentionally memorized were penalized by MiFID. 
This configuration of MiFID significantly reduced the time needed to finish the review, approximately by 5x. 
The results of the manual review is presented in Section~\ref{sec:mem-summary}. 
The review procedure was announced to all participants during\footnote{\url{https://www.kaggle.com/c/generative-dog-images/discussion/106206}} and after\footnote{\url{https://www.kaggle.com/c/generative-dog-images/discussion/102701}} the competition submission period. 

\section{Results and Data Release}
\label{sec:data-release}
A total of $924$ teams joined the competition, producing over $11{,}192$ submissions.
Visual samples from submitted images are shown in the appendix.

\subsection{Data Release}
The complete dataset is released with the publication of this paper to facilitate future work on benchmarking generative modeling\footnote{\url{https://www.kaggle.com/andrewcybai/generative-dog-images}}. 
It includes:
\begin{itemize}
    \item A total of $1{,}675$ submissions selected by users for final scoring, each containing $10{,}000$ generated dog images with dimension $64 \times 64 \times 3$. 
    \item Manual labels for the top 1000 ranked submissions of whether the code is a legitimate generative method and the type of illegitimacy involved if it is not. 
    This was extremely labor-intensive to obtain. 
    \item Crowd-labeled image quality: ~$50{,}000$ human labeled quality and diversity of images generated from the top 100 teams (non-memorized submissions only).
\end{itemize}
 We also release the code to reproduce results in the paper\footnote{\url{https://github.com/jybai/generative-memorization-benchmark}}, as well as demo code to compute MiFID \footnote{\url{https://www.kaggle.com/wendykan/demo-mifid-metric-for-dog-image-generation-comp}}. 

\subsection{Summary of Memorization Study}
\label{sec:mem-summary}
The 1000 top submissions were manually labeled as to whether or not (and how) they intentionally memorized. 
As we predicted prior to the start of the competition, the most pronounced method of cheating was training sample memorization. 
We observed different levels of sophistication in these memorization methods - from very naive (submitting the training images) to highly complex (designing a GAN to memorize).
The labeling results are summarized as follows:

\subsubsection{Memorization GAN (mgan)}
Typical GANs are trained via an iterative method where a generator and a discriminator are updated alternatively to converge towards an equilibrium in a minimax game. 
Memorization GANs are purposely trained to memorize the training set while maintaining the architecture of a typical GAN by modifying the update policy of the generator and discriminator.
The training process is split into two parts.
In the first part, the discriminator is updated by feeding only data from the training set (and nothing from the generator).
The discriminator then degenerates into a classifier of training set membership.
In the second part, the discriminator is frozen and only the generator is updated.
To ensure perfect memorization, one-hot vectors instead of random-sampled vectors are fed into the generator as seeds.
Each one-hot vector will then be mapped to one of the training images.
To avoid mode collapse, hyperparameters are tuned based on conventional GAN metrics such as IS or FID~\cite{memGAN}. 
\subsubsection{Supervised mapping (sup)}
These models directly achieve memorization by training a supervised task mapping distinct inputs (ex. one-hot vectors with dimension the size of the training set) to data from the training set.
This method is easier to identify since the model does not follow the model structure of a typical GAN.
\subsubsection{Autoencoder and VAE Reconstruction (ae)}
Autoencoders trained on the training set can directly be used for reconstruction with high fidelity and slight compression to avoid being punished by the memorization penalty. 
Memorization is achieved trivially by feeding the model with data from the training set.

On the other hand, while variational autoencoders \cite{kingma2013auto,rezende2014stochastic} are legitimate generation methods if the seeds fed into the decoder are randomly sampled, memorization can be achieved by selecting seeds that are encoded training images (passing through the encoder) which are then passed as input to the decoder to basically reconstruct the training set.

\subsubsection{Augmentation (aug)}
These submissions directly generate images by augmenting the training set with typical augmentation techniques such as cropping, morphing, blending and adding noise.
The naivety of this approach makes it the easiest to identify and generally can be filtered out with MiFID.

\subsection{Competition Results Summary}
\begin{figure}
\centering
  \includegraphics[width=0.9\linewidth]{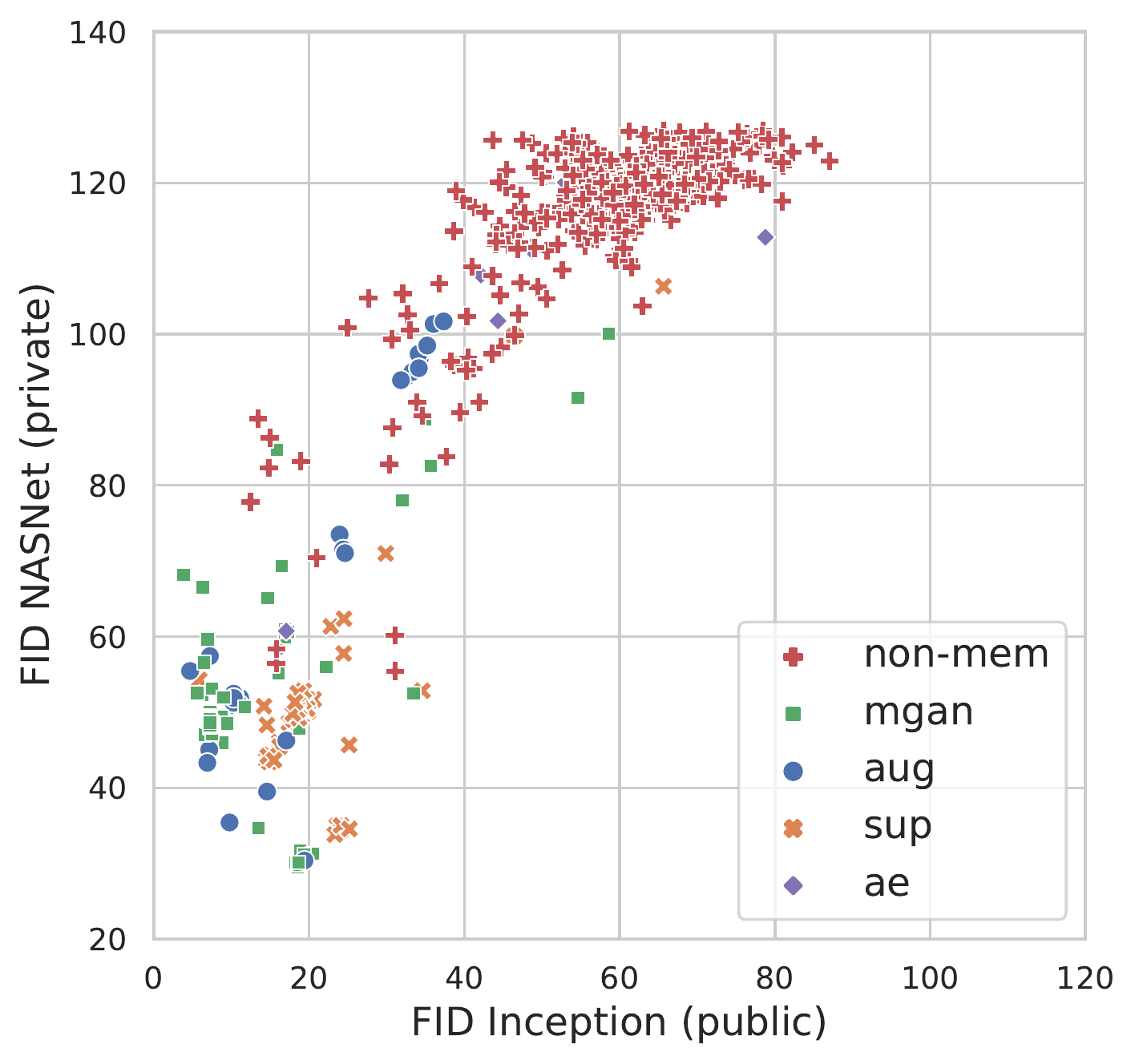}
\caption{Distribution of FID for public vs private scores with manual labels. The better (lower) FIDs are the ones using various memorization techniques.}
\label{fig:labeled-fid}
\end{figure}

\begin{figure}
  \centering
  \includegraphics[width=0.9\linewidth]{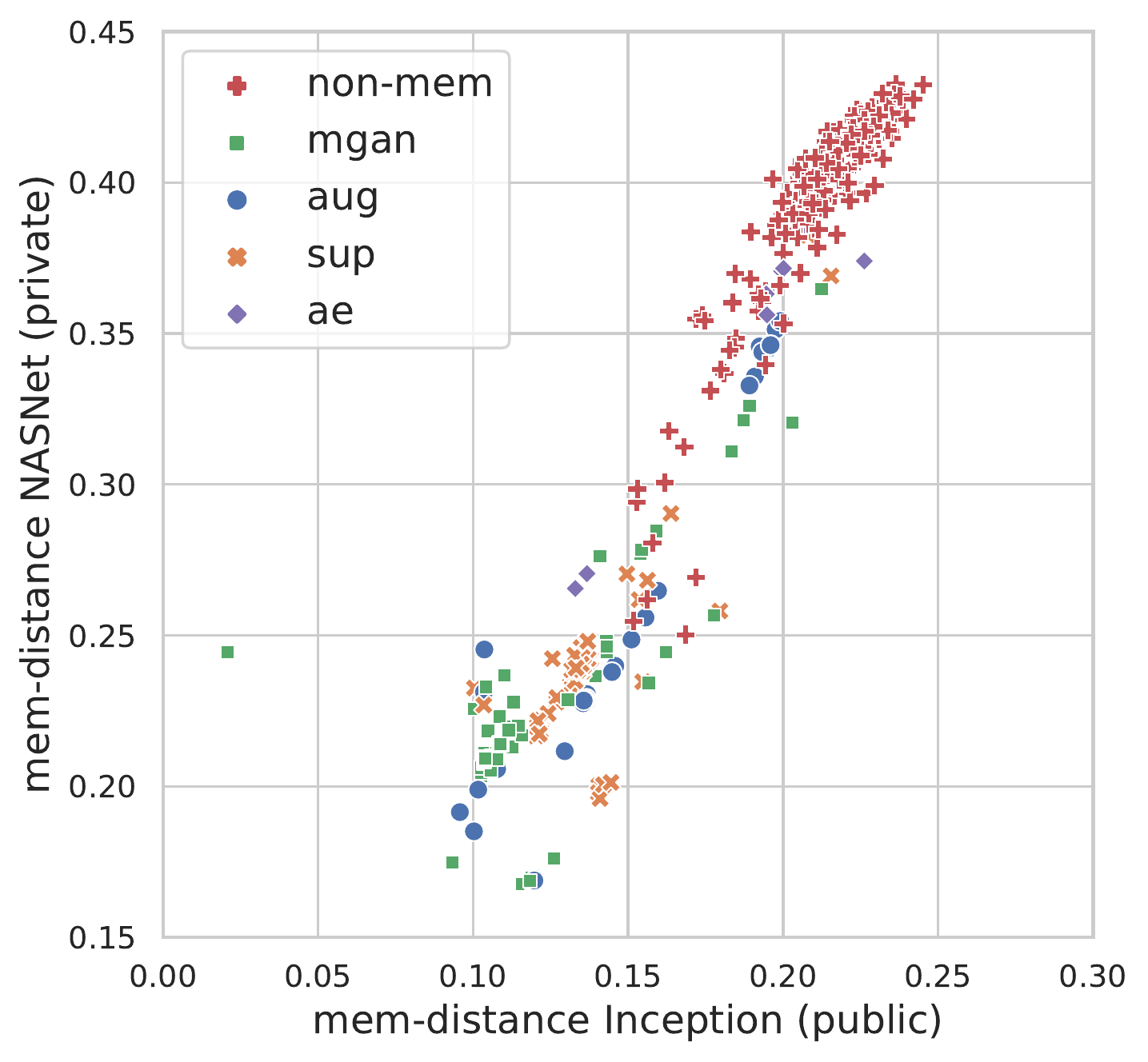}
\caption{Distribution of memorization distances for public vs private scores with manual labels. The better (lower) FIDs are the ones using various memorization techniques.}
\label{fig:labeled-md}
\end{figure}

In Figure~\ref{fig:labeled-fid}, we observe that 
memorizing models 
score extremely good (low) FID scores on both the public and private leaderboard.
Specifically, memorization GAN achieved top-tier performance and it was a highly-debated topic for a long time whether it should be allowed in the competition.
Ultimately, memorization GAN was banned, but it serves as a good reminder that generative-looking models may not actually be generative in the true sense.
In Figure~\ref{fig:labeled-md}, we observe that the range of memorization calculated by NASNet (private) spans twice as wide as Inception (public), allowing easier profiling of intentionally-memorizing submissions by memorization penalty.
It reflects the effectiveness of our strategy selecting the model for calculating private score.

Participants generally started with basic generative models such as DCGAN~\cite{radford2015unsupervised} and moved to more complex ones as they grow familiar with the framework.
Most notably BigGAN~\cite{brock2018large}, SAGAN~\cite{zhang2018selfattention} and StyleGAN~\cite{karras2019style} achieved the most success.
Interestingly, one submission using DCGAN~\cite{radford2015unsupervised} with spectral-normalization~\cite{miyato2018spectral} made it into top 10 in the private leaderboard, suggesting that different variations of GANs with proper tuning might all be able to achieve good FID scores~\cite{lucic2017gans}.

According to the participants' feedback after the competition~\cite{competition_takeaways}, given the computation constraint, fitting a larger model with less epochs, as opposed to a smaller model with more epochs, can help with overfitting and producing more diverse generated instances.
Vanilla batch normalization~\cite{DBLP:journals/corr/IoffeS15} is also not advised, as it causes samples to have higher correlation which results in lower diversity.
They also found that applying simpler models with less hyperparameters are more easily tuned and produced stabler performance.
Finally, participants agree the discrepancy between FID scores calculated with different projections in the public and private leaderboard undermines the belief of FID being projection-space invariant.

\section{Insights}
\label{sec:insights}
\subsection{Unintentional Memorization: models with better FID memorize more}
In our observation, almost all removed submissions intentionally memorize the training set.
This is likely because it is well-known that memorization achieves a good FID score.
The research community has long been aware that memorization can be an issue for FID.
Although several recent studies have proposed methods for detecting memorization in generative modeling~\cite{gulrajani2018towards,meehan2020non}, there are only limited formal studies on benchmarking both the quality and memorization aspects.
This can pose a serious problem when researchers continue to claim state-of-the-art results based on improvements to FID if there is not a systematic way to measure and address training set memorization.
With disturbing findings from our study, we caution the danger of ignoring memorization in research benchmark metrics, especially with unintentional memorization of training data.

\begin{figure}
  \includegraphics[width=0.8\linewidth]{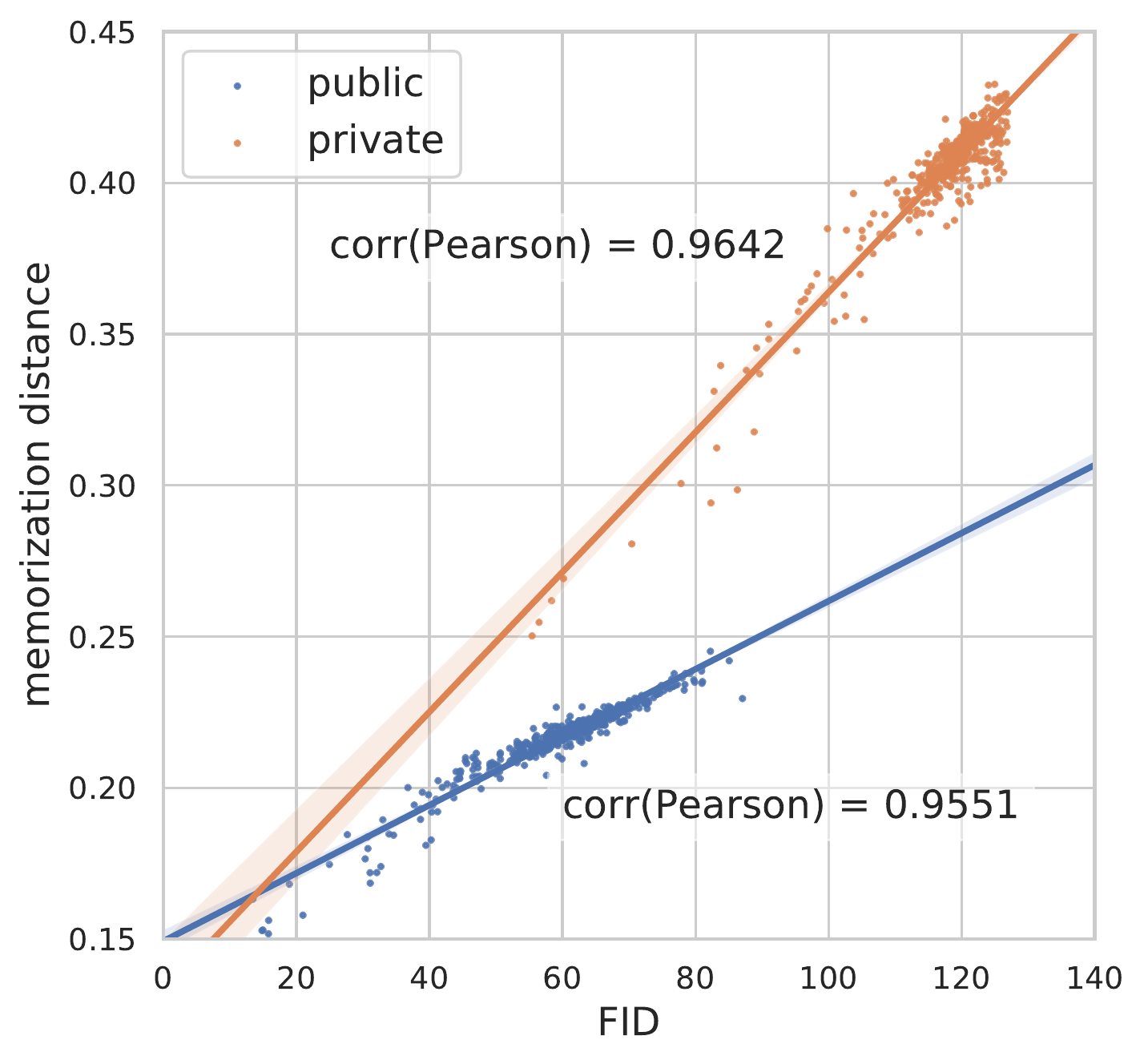}
\caption{FID vs memorization distance distribution with non-memorized submissions. It shows that FID is highly correlated to memorization.}
\label{fig:corrs-fid-md}
\end{figure}

In Figure~\ref{fig:corrs-fid-md}, we plot the relationship between FID and memorization distance for all 500 legitimate models in the public and private leaderboard, respectively.
Note that these models are legitimate, most of which popular variants of state-of-the-art generative models such as DCGAN and SAGAN recently published in top machine learning conferences.
Interestingly and unfortunately, the Pearson correlation between FID and memorization distance is above 0.95 for both leaderboards.
In general, correlation is expected since a generative model should capture the real data distribution which will then generate samples close to the training data.
However, we observe high correlation between FID and memorization distance even in the top 100 submissions where the difference in quality among submissions is barely identifiable with human perception.
In this case, we would expect an unbiased, fair quality metric to be independent from the memorization distance (distance to nearest neighbor in the training set).

\begin{figure}[ht]
  \includegraphics[width=0.8\linewidth]{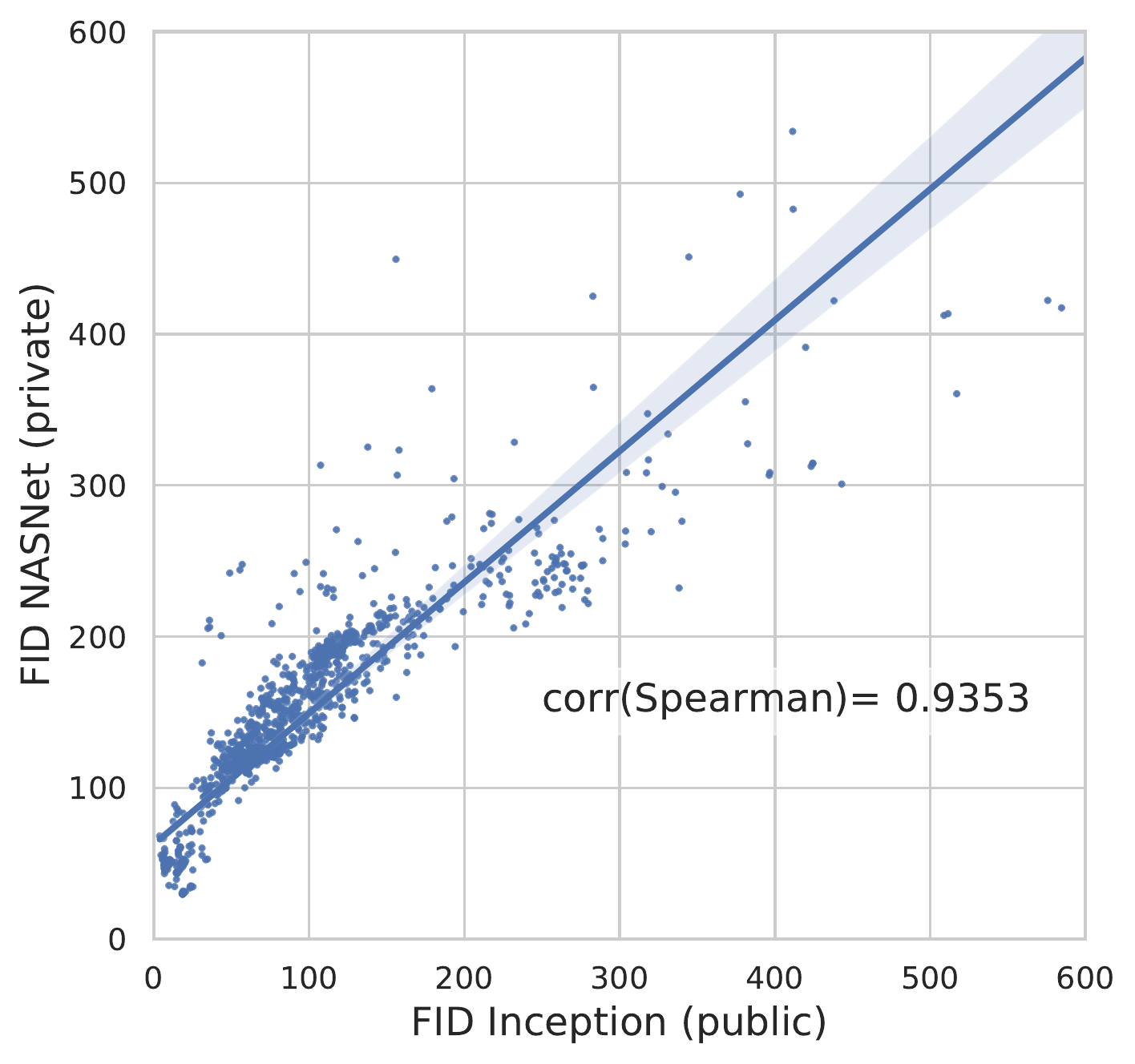}
\caption{Public FID Inception vs private FID NASNet. it shows that FIDs from two pre-trained models are highly correlated.}
\label{fig:corrs-fid}
\end{figure}

We argue that it is important for the research community to take memorization more seriously, given how easy it is for memorization to occur unintentionally.
The research community needs to better study and understand the limitations of current metrics for benchmarking generative models.
When proposing new generative techniques, it is crucial to adopt rigorous inspections of model quality, especially to evaluate training sample memorization. 
Existing methods such as visualizing pairs of generated images and their nearest neighbors in the training dataset should be mandatory in benchmarks. 
Furthermore, other methods such as the FID and memorization distance correlation (Figure~\ref{fig:corrs-fid},~\ref{fig:corrs-fid-md}) for different model parameters can also be helpful to include in publications. 

\subsection{Debunking FID: choice of latent space for feature projection is non-trivial}
In the original paper where FID is proposed~\cite{heusel2017gans}, features from the coding layer of an Inception model are used as the projected latent space to obtain ``vision-relevant'' features.
It is generally assumed that Fr\'echet Inception Distance is invariant to the chosen latent space for projection as long as the space is "information-rich", which is why the arbitrary choice of the Inception model has been widely accepted.
Interestingly, there has not been much study as to whether the assumption holds true even though a relatively large amount of new generative model architectures are being proposed (many of which rely heavily on FID for performance benchmarking).
In our competition, we used different models for the public and private leaderboards in an attempt to avoid models which ``overfit'' to some particular feature space.

In Figure~\ref{fig:corrs-fid}, we examine the relationship between Fr\'echet Distance calculated by two different pre-trained image models that achieved close to state-of-the-art performance on ImageNet classification (specifically, Inception~\cite{szegedy2016rethinking} and NasNet~\cite{zoph2016neural}).
At first glance, a Spearman correlation of 0.93 seems to support the assumption of FID being invariant to the projection space.
However, on closer inspection we noticed that the mean absolute rank difference is 124.6 between public and private leaderboards for all 1675 effective submissions.
If we take out the consistency of rank contributed by illegitimate submissions by considering the top 500 labeled, legitimate submissions only, the mean absolute rank difference is as large as 94.7 (18.9~\%).
To put it into perspective, only the top 5 places receive monetary awards and there is only 1 common member between the top 5 evaluated by FID projected with the two models.

It is common to see publications claiming state-of-art performance with less than $5\%$ improvement compared to others.
As summarized in the Introduction section of this paper, generative model evaluation, compared to other well-studied tasks such as classification, is extremely difficult.
Observing that model performance measured by FID fluctuates in such great amplitude compared to the improvements claimed by many newly-proposed generative modeling techniques, we would suggest taking progression on the FID metric with a grain of salt.

\section{Related work}
Our competition results verified a well-known issue with generative modeling metrics - training sample memorization is highly correlated to the most popular benchmark metrics, IS and FID~\citep{borji2019pros}.
Although this is a well-known issue, most studies in generative modeling have chosen to believe that this only happens in extreme or hypothetical settings.
On the other hand, there are alternative metrics shown to be more sensitive to memorization.
Borji et al. reported that the Average Log Likelihood~\cite{goodfellow2014generative,theis2015note}, Classifier Two-sample Tests~\cite{lehmann2006testing}, and Wasserstein Critic~\cite{arjovsky2017wasserstein} are capable of detecting some level of memorization.
However, the lack of awareness regarding the severity of memorization and the popularity of the Inception Score~\citep[IS;][]{salimans2016improved} and Fr\'echet Inception Distance~\citep[FID;][]{heusel2017gans} have hindered the adoption or development of better generative modeling benchmark metrics.
Recent studies have proposed metrics to address model generalization in the perspectives of data-copying~\cite{gulrajani2018towards, meehan2020non} and cross-domain consistency (text, sound, images)~\cite{grnarova2018domain}.
However, there is still no comprehensive metric that takes both generation quality and memorization into account.

We believe that a good benchmark metric should be extensively tested for its various properties and potential limits before being widely adopted, making public competitions the perfect testing ground.
Our work contributes to the field of generative modeling by showcasing the feasibility of generative modeling competitions and providing the $11{,}000+$ collected submission results in our competition as an standard benchmark dataset for benchmark metrics.

\section{Conclusions}
We summarized our design of the first ever generative modeling competition and shared insights obtained regarding FID as a generative modeling benchmark metric.
By running a public generative modeling competition, we observed how participants attempted to game the FID, specifically with memorization, when incentivized with monetary awards.
Our proposed Memorization-Informed Fr\'echet Inception Distance (MiFID) effectively punished models that intentionally memorize the training set which current popular generative modeling metrics do not take into consideration.
We are not suggesting that MiFID is a drop-in replacement for FID in general but rather an efficient profiling tool suitable for competition settings.

We shared two main insights from analyzing the $11{,}000+$ submissions.
First, unintentional training sample memorization is a serious and possibly widespread issue.
Careful inspection of the models and analysis on memorization should be mandatory when proposing new generative model techniques.
Second, contrary to popular belief, the choice of pre-trained model latent space when calculating FID is non-trivial.
The top 500 labeled, non-memorized submission mean absolute rank difference percentage between our two models is 18.9~\%, suggesting that FID is rather unstable to serve as the benchmark metric for new studies to claim minor improvement over past methods.

\begin{acks}
The authors would like to thank Phil Culliton for the contribution in metric development; Douglas Sterlin for code-only competition platform development and support; Julia Elliott for business support and human labeling work. Bai and Lin were partially supported by the MOST of Taiwan under 107-2628-E-002-008-MY3.
\end{acks}

\bibliographystyle{ACM-Reference-Format}
\bibliography{kaggle-gan}

\appendix
\section{Generated Samples Visualization}
\begin{figure}[ht]
\centering
  \includegraphics[width=0.9\linewidth]{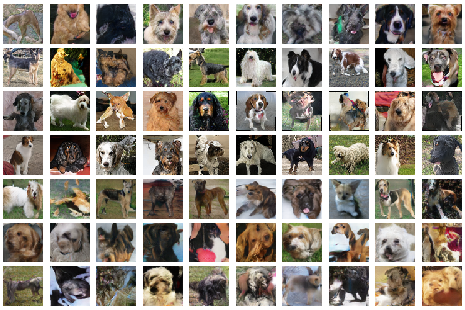}
  \caption{Submissions from ranks 1 (first row), 2, 3, 5, 10, 50, 100 (last row) on the private leaderboard. Each row is a random sample of 10 images from the same team. Visually, the quality of the generated images gets lower as the ranks get higher. }
  \label{fig:dogs}
\end{figure} 
\end{document}